\documentclass[letterpaper, 10 pt, conference]{ieeeconf}

\usepackage{times}  
\usepackage{helvet} 
\usepackage{courier}  
\usepackage[hyphens]{url}  
\usepackage{graphicx} 
\usepackage{xcolor}

\usepackage{soul}
\usepackage[utf8]{inputenc}
\usepackage{booktabs}
\usepackage{algorithmic}
\usepackage{mathtools}
\usepackage{amsmath} 
\usepackage{amssymb}  
\usepackage{multirow}

\usepackage[ruled,noend,linesnumbered]{algorithm2e}
\usepackage{standalone}
\usepackage{amsfonts}
\usepackage[mathscr]{euscript}
 \let\mathscr\relax

\usepackage{accents}

\usepackage{stackengine}

\let\labelindent\relax

\usepackage{enumitem}

\SetAlFnt{\footnotesize}
\SetAlCapFnt{\footnotesize}
\SetAlCapNameFnt{\footnotesize}

\SetCommentSty{mycommfont}

\setlist{  
  listparindent=\parindent,
  parsep=0pt,
}

\title{\LARGE \bf Waiting Tables as a Robot Planning Problem}

\author{Anahita Mohseni-Kabir$^{1}$, Manuela Veloso$^{1}$, and Maxim Likhachev$^{1}$

\thanks{*This work was partially supported by Sony AI.}
\thanks{$^{1}$The authors are with School of Computer Science, Carnegie Mellon University {\tt\small \{anahitam, mmv, maxim\}@cs.cmu.edu}.}
}

\IEEEoverridecommandlockouts

\begin{document}

\maketitle
\thispagestyle{empty}
\pagestyle{empty}

\begin{abstract}
We present how we formalize the waiting tables task in a restaurant as a robot planning problem. This formalization was used to test our recently developed algorithms that allow for optimal planning for achieving multiple independent tasks that are partially observable and evolve over time~\cite{mohseni2020efficient,mohseni2021technical}.


\end{abstract}

\section{Introduction}

\noindent A robot waitress working in a restaurant should take care of an ongoing stream of tasks, including delivering hot food, taking drink orders, taking food orders, checking on customers who have just received their meal, and finally, cashing out a table about to leave. Attending the customers' needs in a timely and efficient manner is very challenging. One approach to address this problem would be to attend the customers on a first come, first served basis, \emph{i.e.}, based on the amount of time each customer was waiting. This approach might be practical in some domains, but is not effective in a restaurant domain where the customers' satisfaction does not only depend on the wait time, but it also depends on the task's features, and how important the task is. In this work, we explain how we formalize the dining process as a robot planning problem.

The restaurant domain has the following two main challenges. First, the robot waitress should be able to effectively multitask by keeping track of everything that needs to be done and prioritizing them so the most important tasks get done quickly. For example, a robot who has just received an order from a table should send that order to the kitchen as soon as possible so things don't get backed up down the line. Second, the robot waitress should plan routes around the restaurant, taking into account the tasks that needs to be done at each station. The fewer trips the robot makes between the bar and the kitchen, the sooner the customers get their food. For example, if the robot needs to pass a table of customers on the way to the bar, it should stop by the table briefly to check on the customers before proceeding to the bar. 

Restaurant waiters are capable of addressing these two challenges (and many more challenges) by 1) leveraging an internal model of how the customers needs and satisfaction evolve throughout the dining process, and 2) multitasking and navigating in the restaurant to serve as many people as possible. By using these models the waiter is able to make decisions on what actions she should perform to gain the maximum final tip. We model the waiting table task as a planning problem to enable a robot to address the same challenges as a restaurant waiter. 

There has been work on robot localization and navigation in a restaurant~\cite{yu2012autonomous,qing2010research}. Different from these works, we assume the robot is capable of localizing within the restaurant, and we focus on how the customers' satisfaction of the service, which is not directly observable, is affected by the robot's actions. Task planning is another area of research that has been studied in the service robot domain~\cite{na2015robot,jo2010task}. These works either do not model the customers' satisfaction or model it as an observable variable and use it to prioritize the next task. Differently, we are interested in how the robot's sequence of actions maximizes the customers' long-term satisfaction. Furthermore, these works consider each task as an indivisible task. We allow the robot to diverge from its current task to service more customers and make fewer trips. There has been work on predicting customer's state in a restaurant or a bar~\cite{chen2018information}. They focus on inferring the customers internal state and using that to select a robot behavior. In contrast, we focus on how the robot's sequence of actions impacts the customers' long-term satisfaction. %

We enable the robot to reason about the unknown customers' satisfaction so that it can effectively prioritize the tasks to keep all the customers satisfied. The robot should also reason about its long-term effects of immediate actions to foresee what will happen in the future so it can plan ahead. Partially observable markov decision processes (POMDPs)~\cite{cassandra1998survey} provide a mathematical tool to achieve these objectives. A POMDP is capable of modeling a robot's sequential decision making process, while also being able to represent uncertainty in the robot's execution and perception. POMDPs have been applied to many real-world problems in the context of human-robot interaction~\cite{bai2015intention,nikolaidis2017human}. However, we are not aware of any approaches that leverage multiple POMDP models to formalize the restaurant domain.

\vspace{-0.1cm}

\section{Background on POMDPs}
\noindent A POMDP is a framework for modeling an agent's decision process under uncertainty. Formally, a POMDP is a tuple $(S, A, Z, T, O, R, \gamma)$, where $S$ denotes the agent's state space, $A$ denotes the agent's action space, and $Z$ denotes the agent's observation space. At each time step, the agent takes an action $a \in A$ and transitions from a state $s \in S$ to $s' \in S$ with probability $T (s, a, s')=p(s'|s,a)$, makes an observation $z \in Z$, and receives a reward equal to $R(s,a)$. The conditional probability function $O(s' , a, z) = p(z|s',a)$ models noisy sensor observations. The discount factor $\gamma$ specifies how much immediate reward is favored over more distant reward. The objective is to choose actions at each time step to maximize its expected future discounted reward: $E\left[\sum _{{t=0}}^{\infty }\gamma ^{t}r_{t}\right]$,  where $r_{t}$ is the reward gained at time $t$.

\section{Restaurant Formulation}

\begin{figure}[tb]
\centering
\includegraphics[width=0.95\columnwidth]{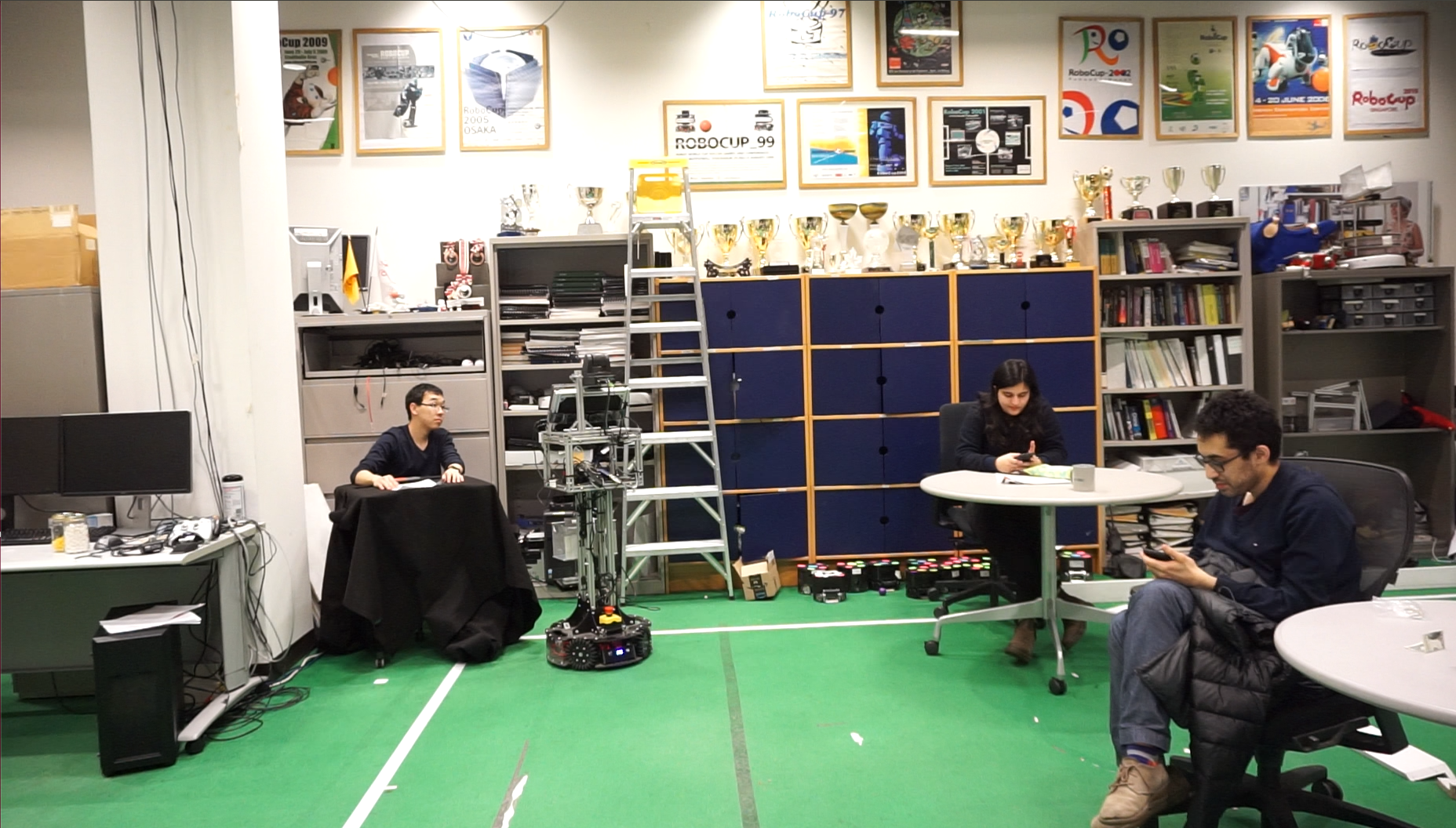}
\caption{A restaurant setting with 3 tables and one robot.} 
\vspace{-0.5cm}
\label{img:restaurant}
\end{figure}


\noindent As illustrated in Fig.~\ref{img:restaurant}, we consider a restaurant setting where one robot services $N$ tables. We do not model each customer in the restaurant individually and only assume one model for each table in the restaurant. We assume that the states of the humans on a particular table are aggregated to have one estimate for the table.
We formalize the waiting tables for a restaurant task as a planning problem with one robot and $N$ independent models for the $N$ tables in the restaurant. We consider each table from when the customers sit at the table to when they leave as one task. The robot keeps a distribution over the state of the tasks and updates them as it executes actions. At each time step, the robot decides what action should be executed with respect to which task.This enables the robot to consider switching between the tasks after each action execution.
The robot can only execute one action at each time that depends on the duration  of  possible  actions,  the  state  of  each  table,  and  how these tables evolve, \emph{e.g.}, the table becomes unsatisfied if it is not served soon or the food becomes cold after a few time steps. While the robot services one table, the other tables evolve according to their underlying Hidden Markov Model (HMM). 

\begin{figure}
\centering
\includegraphics[width=1\columnwidth]{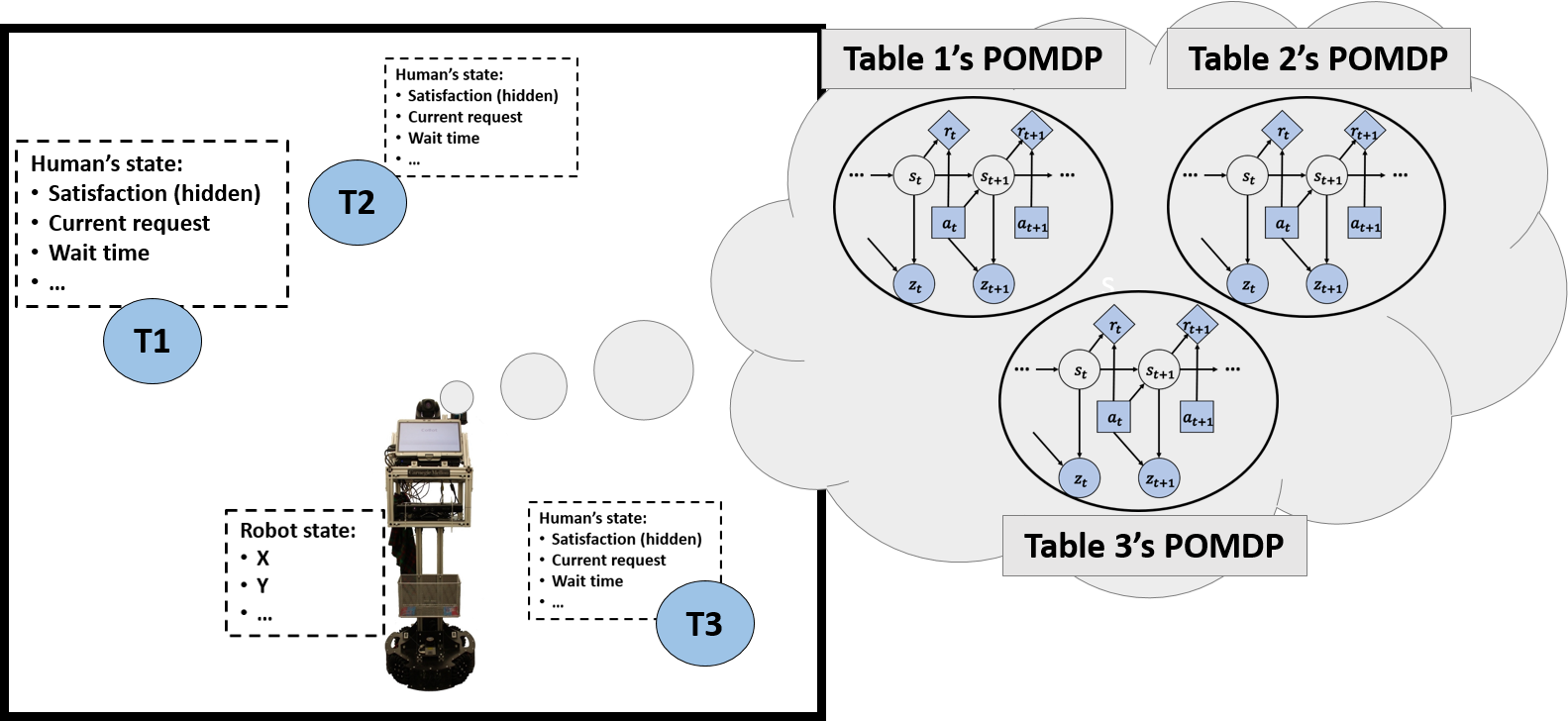}
\vspace{-0.6cm}
\caption{The robot operates in a restaurant with $3$ tables, \textit{T1}, \textit{T2}, and \textit{T3}. We model each table, the state of the robot and the actions that can be applied to this particular table as a POMDP. The robot has $N$ POMDP models for a restaurant with $N$ tables.} 
\vspace{-0.5cm}
\label{img:pomdp_models}
\end{figure}

To enable the robot to perform the waiting tables task, we model each table, the state of the robot and the actions that can be applied to this particular table as a POMDP. Thus, there are $N$ POMDPs in the restaurant that share the robot between themselves as shown in Fig.~\ref{img:pomdp_models}. This POMDP representation enables the robot to reason about the uncertainty in the table's internal state and its own actions and how the dining process for that specific table evolves after a sequence of action executions.

We provide the general representation of the POMDP model for a table. We also discuss an example implementation of each element of the POMDP model for the restaurant setting that we used in the experiments of~\cite{mohseni2020efficient} and~\cite{mohseni2021technical}.

\begin{itemize}
    \item \textbf{State space $S$}: For one table, the full state of its POMDP is $S = SR \times SC$ where $SR$ is the robot's state space and $SC$ is the human's state space. If we represent an element of the robot's state $sr \in SR$ and an element of the human's state $sc \in SC$ in vector form, an element of the table's state $s \in S$ is a concatenation of the robot's and human's state $s=(sr,sc)$. The robot's state variables include robot's state information that will be shared between the tables, and it does not include any human specific information. The rest of the state variables $sc$ are specific to each table. 
    
    \begin{itemize}
        \item Robot's state variables $sr \in SR$ (\emph{e.g.}, \textit{position})
        \item Human's state variables $sc \in SC$ (\emph{e.g.}, \textit{satisfaction level}, \textit{wait time}, \textit{cooking status} and  \textit{current request})
    \end{itemize}
    \medbreak
   \noindent \textbf{Example $S$}: We enumerate what state variables we use with their range in brackets in front of each variable. The state variables can take any integer values from the first number in the bracket to the last number inclusive. We also mention what each integer value represents. The time related variables have values from $0$ to $time_{max} = \#\text{ tables} \times {sat_{max}}$ where ${sat_{max}}$ is the highest value for the \textit{satisfaction} variable. This accounts for having more time to service the customers when there are more tables. 
    
    \begin{itemize}
        \item Robot's state $sr$ contains \textit{x} and \textit{y} ($11 \times 11$ grid). 
        \item Human's state $sc$ contains 
            \begin{itemize}
                \item {Satisfaction} $[0,5]$: very unsatisfied ($0$), unsatisfied ($1$), slightly unsatisfied ($2$), neutral ($3$), satisfied ($4$), very satisfied ($sat_{max} = 5$)
                \item {Food} $[0,3]$: not served ($0$), plate-full ($1$), plate-half ($2$), plate-empty ($3$)
                \item {Water} $[0,3]$: not served ($0$), glass-full ($1$), glass-half ($2$), glass-empty ($3$)
                \item {Cooking status} $[0,2]$: cooking-started ($0$), food-half-ready ($1$), food-ready ($2$)
                \item {Current request} $[1,8]$: want-menu ($1$), ready-to-order ($2$), want-food ($3$), want-drinks ($4$), want-bill ($5$), cash-ready ($6$), cash-collected ($7$), table-needs-to-be-cleaned ($8$)
                \item {Hand raise} $[0,1]$: no-hand-raise ($0$), hand-raise ($1$). This variable represents if the customers have a request or not. This variable is always $1$. After the customers leave, it becomes $0$.
                \item {Time since food or water has been served} $[0,time_{max}]$: this variable represents the number of time steps since the customers started eating or drinking. It affects the value of \textit{food} and \textit{water}. We provide more details below.
                \item {Time since food is ready} $[0,time_{max}]$: this variable represents for how many time steps the food (or drinks) has been ready to be delivered to the customers. We provide more details below.
                \item {Time since request} $[0,time_{max}]$: this variable represents for how many time steps the customers at the table have been waiting to be serviced. 
                
            \end{itemize}
    \end{itemize}
    
    \bigbreak
    \item \textbf{Action space $A$}: The full action space is a set with all the following actions $A=\{AN,AC,AS,AI,\textit{no op}\}$.
    \begin{itemize}
        \item Navigation actions $AN$ (\emph{e.g.}, \textit{go to the table} and \textit{go to the kitchen}) 
        \item Communication actions $AC$ ( \emph{e.g.}, \textit{food is not ready} and \textit{I'll be back to serivce your table})
        \item Service actions $AS$ (\emph{e.g.}, \textit{serve food} and \textit{give the bill})
        \item Information gathering actions $AI$ (\emph{e.g.}, \textit{ask if the customers are ready for the bill})
        \item A special \textit{no operation} action
    \end{itemize}
      \medbreak
     \noindent \textbf{Example $A$}: $A=\{AN,AC,AS,\text{no op}\}$.
    \begin{itemize}
        \item Navigation actions $AN$: \textit{go to the table}.        
        \item Communication actions $AC$: \textit{food is not ready} and \textit{I'll be back to your table later}.
        \item Service actions $AS$: One \textit{serve} action.  Depending on the table's \textit{current request}, the appropriate service action gets executed. For example, if the table's \textit{current request} is \textit{want-menu}, executing the \textit{serve} action when the robot is at the table represents handing over the menu.
        \item The special \textit{no operation} action.
    \end{itemize}
    
    \noindent In this instance of the restaurant model, we do not include information gathering actions. 
    \bigbreak
    \item \textbf{Transition function $T$}: We assume the robot's next state $sr'$ is independent of the human's current and next states, $sc$ and $sc'$, and only depends on its own current state $sr$ and action $a$. Similarly, the human's next state $sc'$ is independent of the robot's current and next states, $sr$ and $sr'$, and only depends on human's state $sc$ and action $a$. Thus, we can decouple the transition function as follows $T (s, a, s')=\Pr(sc'|sc,a)\Pr(sr'|sr,a)$. 
    \medbreak
    \noindent \textbf{Example $T$}:
    We assume that the robot's actions with respect to its own state are deterministic (\emph{i.e.}, robot's position changes deterministically). Each table goes through a consecutive sequence of $8$ requests that we mentioned above (the $8$ values for the \textit{current request} variable). The table gets served if the robot performs the \textit{serve} action at the table. After each \textit{serve} action, the value of the \textit{current request} is updated to the next value (\textit{current request}$+1$).  The \textit{time since request} variable resets to $0$ when a table is served, otherwise it goes up till it reaches its maximum value. The actions increase the relevant time-related variables of the state by $1$ except the navigation actions which can take $1$, $2$, or $3$ time steps depending on where the robot is with respect to the tables. 
    
    If a table is waiting for the food, the \textit{cooking status} variable increases by $1$ after $\frac{time_{max}}{3}$ time steps until the food is fully cooked and ready to be served, \emph{i.e.}, \textit{cooking status} $=2$. 
    This means the kitchen takes $\frac{2*time_{max}}{3}$ time steps to prepare the food.
    When the food is ready to be served, the \textit{time since food is ready} variable keeps increasing until the robot serves the table, \emph{i.e.}, the robot goes to the table and executes the \textit{serve} action, or it reaches its maximum value. If the value of the \textit{current request} is \textit{want-food} or \textit{want-drinks} and the robot serves food or drinks, the table starts its eating or drinking process. 
    We assume each table takes $\frac{2*time_{max}}{3}$ time steps to finish eating (or drinking).
    The \textit{time since request} variable does not go up when the people are eating or drinking. 
    The \textit{time since served} variable resets to $0$ when the food or drinks are served and goes up till the customers finish their food ($food=3$) or water ($water=3$). 
   The \textit{time since served} variable is used to keep track of time while the customers are eating or drinking and affects the value of \textit{food} and \textit{water}. The \textit{food} and \textit{water} variables represent the table's eating and drinking process and increase after $\frac{time_{max}}{3}$ time steps until they reach their maximum value. 
    
    All the variables except \textit{satisfaction} transition deterministically. The \textit{satisfaction} variable goes down by $1$ after $\frac{time_{max}}{sat_{max}}$ time steps; if a customer is very satisfied at the beginning and does not get served within $time_{max}$, she becomes very unsatisfied. When the table is waiting for food, the \textit{satisfaction} variable goes down by $1$ after $\frac{time_{max}}{sat_{max}+1}$ time steps. Here the assumption is that if the people are waiting for food, their satisfaction level decreases faster than when they are waiting for other reasons. 
    When a table is served its \textit{satisfaction level} changes stochastically. If they are very unsatisfied, their satisfaction level increases by $1$, $30\%$ of the times, and stays the same, $70\%$ of the times. If they are very satisfied, their satisfaction level does not change. Otherwise, their satisfaction level increases by $1$, $60\%$ of the times, and stays the same, $40\%$ of the times. This means that it is much harder to make the very unsatisfied tables a bit more satisfied than making the satisfied tables very satisfied. 
    
    \bigbreak
    \item \textbf{Observation space $Z$}: We assume that the robot's state is fully observable. 
    The human's state $sc$ may be partially observable. For example, the robot may execute an action and observe the human's \textit{neediness} which might give information regarding the human's \textit{satisfaction level}. For the part of the human's state that is observable, the model has the same number of observation variables as the number of state variables. The model can have any number of observation variables for the partially observable part of the state $sc$.
    \medbreak
    \noindent \textbf{Example $Z$}:
     In an instance of the restaurant model that we consider, we assume that we have one observation variable per each state variable except for the \textit{satisfaction level} that is hidden. As the robot's state is fully observable and the actions of the robot with respect to its own state are deterministic, we do not consider any observations associated with the robot's state.
    This means we have $8$ observation variables for the following state variables: {food}, {water}, {cooking status}, {current request}, {hand raise}, {time since food or water has been served}, {time since food is ready}, and {time since request}. 
    

    \bigbreak
    \item \textbf{Observation function $O$}: 
    The model does not have observations for the robot's state. The human's state can be partially observable, thus the observation function only depends on the human's state $sc'$, action $a$ and the observation of the human's state $z$, $O(s',a,z) = \Pr(z|sc',a)$. 
    \medbreak
    \noindent \textbf{Example $O$}:
    We only consider \textit{satisfaction level}, $sat$, as a hidden variable, and the other variables, let's call them $zh$, are observable. This means that $\Pr(z_{t+1}=zh|sat,zh,a)=1$ and $\Pr(z_{t+1}=zh'|sat,zh,a)=0$ where $zh' \neq zh$. Since \textit{satisfaction level} is the only hidden variable, the belief state of the POMDP keeps a distribution over this variable.
    
    \bigbreak
    \item \textbf{Reward function} \textit{R}: The reward function specifies how the robot should service the table. 
    \medbreak
    \noindent \textbf{Example $R$}:
    In an instance of the restaurant model that we consider, we consider servicing a table has a positive reward inversely proportional to the table's satisfaction. If the table is unsatisfied and waiting to be served, a negative reward is given. Navigation actions incur a negative reward. This reward function encourages the robot to service the table sooner if it is unsatisfied. The reward function is as follows:

\small  \begin{equation*} \label{eq:rew_sat}
\begin{aligned}
&time = min(\text{time since request},10) \\
&R(s,\text{serve}) = 5 * ({sat}_{max} - sat'+1) \\ &R(s,\text{go to}) = \frac{-\text{distance to table}}{3}\\
&R(s,\text{other actions}) =
\begin{cases} 
-2^{time} & sat' = 0 \\ 
-1.7^{time} & sat' = 1 \\ 
-1.4^{time} & sat' = 2 \\
1 & sat' = 3,4,5; sat'>sat \\
0 & \text{otherwise}
\end{cases} 
\end{aligned}
\end{equation*}  \normalsize
    
\end{itemize}

\noindent This concludes the representation of the POMDP model for a single table. 
In the next section, we provide a discussion on what assumptions we made regarding the restaurant model and how that would relate to a real restaurant setting. We add a subscript $i$ to the elements of table $i$'s POMDP model when we refer to them. 


\section{Discussion}
\noindent We assume each table goes through the dining process independently of each other. More specifically, our formulation makes the following assumptions regarding the POMDPs:

\begin{itemize}
    \item We make the following assumptions regarding the state, action and observation space of the tables. The  models' state space only share the robot's state $sr$ between themselves. The rest of the state variables $sc$ are specific to each table. Other than the \textit{no operation} action, the  models' action space do not share any other actions. The models' observation space do not share any observation variables. We believe that these assumptions can be valid in real-world restaurants unless the customers on two different tables booked a reservation together and are going through the dining process together, in which case they should be considered as one table. 
    
    \item Table $i$'s transition probabilities are a function of the human's ($sc,sc' \in SC_i$) and robot's current and next state ($sr,sr' \in SR$), and the robot's action ($a \in A_i$). The transition probabilities do not depend on the state of the other tables. When an action gets executed by the robot, if the action belongs to table $i$, $a \in A_i$, all other tables other than table $i$ transition as if \textit{no operation} has been executed on them for the duration of $a$.
    
    In reality this assumption might be invalid, and $T_i$ might depend on the state of table $j$ or the specific action that is being executed on table $j$ ($a \in A_j$). For example, if table $i$ perceives that the customers at table $j$ that have been waiting less than them got served before them, their level of satisfaction might decrease. 
    
    \item Table $i$'s observation probabilities are a function of the human's state ($sc \in SC_i$), the robot's action ($a \in A_i$), and the robot's observations of the table $i$ ($z \in Z_i$). The observation function does not depend on the state or robot's observations of the other tables.
    
    In reality this assumption might be invalid, \emph{e.g.}, if a table's observed neediness changes based on a nearby table's status, \emph{e.g.}, if the nearby table is eating.
    
    \item The tip or the reward that the robot receives from a table only depends on the human's ($sc,sc' \in SC_i$) and robot's current and next state ($sr,sr' \in SR$), and the robot's action ($a \in A_i$). The total tip or the reward the robot gets is a sum of all the tips or rewards from all the tables in the restaurant.
    
    In reality this assumption might be invalid. For example, if the reward that is given to the robot is based on being served after a nearby table.
    
\end{itemize}


\noindent Although some of the assumptions that we made regarding the restaurant domain might not always hold in a real-world restaurant setting, we believe they provide a good approximation of a real-world restaurant setting. Relevant works on the restaurant domain also implicitly make these assumptions. Our work extends the previous works by looking at the outcome of robot decisions in the long-run and modeling the internal state of the humans.

To perform the waiting tables task, the $N$ POMDP models for the $N$ tables are combined
into one large POMDP model for the whole restaurant. This large POMDP model includes all the tables’ states, actions and observations. The robot then solves this large combined model to decide what action should be executed next on which task. In~\cite{mohseni2020efficient,mohseni2021technical}, we provide algorithms that exploit the independence between the tasks to efficiently solve this large combined model. 

\bibliographystyle{IEEEtran}

\bibliography{main}

\end{document}